\documentclass[sigconf]{acmart}

\usepackage{algorithmic}
\usepackage{algorithm}
\usepackage{multirow}
\usepackage{hyperref}
\usepackage[caption=false]{subfig}
\AtBeginDocument{%
  \providecommand\BibTeX{{%
    \normalfont B\kern-0.5em{\scshape i\kern-0.25em b}\kern-0.8em\TeX}}}

\copyrightyear{2021} 
\acmYear{2021} 
\setcopyright{acmcopyright}\acmConference[KDD '21]{Proceedings of the 27th ACM SIGKDD Conference on Knowledge Discovery and Data Mining}{August 14--18, 2021}{Virtual Event, Singapore}
\acmBooktitle{Proceedings of the 27th ACM SIGKDD Conference on Knowledge Discovery and Data Mining (KDD '21), August 14--18, 2021, Virtual Event, Singapore}
\acmPrice{15.00}
\acmDOI{10.1145/3447548.3467137}
\acmISBN{978-1-4503-8332-5/21/08}

\begin{document}
\fancyhead{}
\title{Time Series Anomaly Detection for Cyber-physical Systems via Neural System Identification and Bayesian Filtering}

\author{Cheng Feng}
\email{cheng.feng@siemens.com}
\affiliation{%
  \institution{Siemens AG}
  \city{Beijing}
  \country{China}
}

\author{Pengwei Tian}
\email{pengwei.tian@siemens.com}
\affiliation{%
  \institution{Siemens AG}
  \city{Beijing}
  \country{China}
}

\begin{abstract}
Recent advances in AIoT technologies have led to an increasing popularity of utilizing machine learning algorithms to detect operational failures for cyber-physical systems (CPS). In its basic form, an anomaly detection module monitors the sensor measurements and actuator states from the physical plant, and detects anomalies in these measurements to identify abnormal operation status. Nevertheless, building effective anomaly detection models for CPS is rather challenging as the model has to accurately detect anomalies in presence of highly complicated system dynamics and unknown amount of sensor noise. In this work, we propose a novel time series anomaly detection method called Neural System Identification and Bayesian Filtering (NSIBF) in which a specially crafted neural network architecture is posed for system identification, i.e., capturing the dynamics of CPS in a dynamical state-space model; then a Bayesian filtering algorithm is naturally applied on top of the ``identified" state-space model for robust anomaly detection by tracking the uncertainty of the hidden state of the system recursively over time. We provide qualitative as well as quantitative experiments with the proposed method on a synthetic and three real-world CPS datasets, showing that NSIBF compares favorably to the state-of-the-art methods with considerable improvements on anomaly detection in CPS.
\end{abstract}




\maketitle

\section{Introduction}
\label{sec:introduction}
Cyber-physical systems (CPS) are commonly used for the monitoring and controlling of industrial processes in important infrastructure assets such as power plants, gas pipelines and water treatment facilities. The active monitoring of the sensor readings and actuator states in such systems for early detection of unexpected system behavior is highly critical as it allows for timely mitigating actions – such as fault checking, predictive maintenance and system shutdown – to be taken such that potential economical and environmental loss caused by system failures can be avoided. However, with the increasing complexity of modern CPS, traditional anomaly detection mechanisms based on manually crafted evolution rules in CPS becomes insufficient. As a result, utilizing machine learning algorithms to build data-driven anomaly detection frameworks in CPS has become a clear tendency thanks to the recent advances in AIoT (AI+IoT) techniques. 

Detecting anomalous behavior on a set of correlated time series signals has been an active research area in the machine learning community for a long time \cite{chandola2009anomaly,chalapathy2019deep}. Many approaches have been proposed for this multivariate time series anomaly detection task. For example, the traditional unsupervised anomaly detection algorithms such as one-class SVM \cite{manevitz2001one} and isolation forest \cite{liu2008isolation} can be applied on the CPS data with time-series patterns ignored or implicitly captured. A more commonly used method is the residual-error based anomaly detection. Specifically, the residual-error based anomaly detection either relies on a predictive model such as recurrent neural networks \cite{hochreiter1997long} to predict future sensor measurements, or a reconstruction model such as autoencoders \cite{lecun2015deep} to compress sensor measurements to lower dimensional embeddings and reconstruct them afterwards. Then the predicted or reconstructed measurements are compared with observed measurements to generate a residual error. An anomaly is detected if the residual error exceeds a predefined threshold. In practice, a robust threshold on residual errors of sensor measurements in CPS is however hardly definable because of the uncertainties from various sources such as sensor noise caused by imperfect manufacturing and process noise caused by imperfect control. As a result, although such methods have shown superior performance on many conventional time series anomaly detection tasks \cite{malhotra2016lstm,chen2017outlier,park2018multimodal,kieu2019outlier,zhang2019deep,audibert2020usad}, their performance may deteriorate on the CPS data where the signals contain unstable noises over time. Density-based anomaly detection can mitigate this issue by modelling the likelihood of observed measurements. For example, state estimation algorithms \cite{ding2020secure} that utilize state-space models and the Kalman filter \cite{chui2017kalman} to track the uncertainty of the true system states under noisy measurements are commonly used for CPS due to their robustness to sensor and process noises. However, as CPS become more ubiquitous and commoditized with the trend of IoT, the requirement of heavy domain knowledge and lack of generality often restrict their applicability in practice. Recent work on density-based anomaly detection based on deep learning and variational methods \cite{an2015variational,xu2018unsupervised,zong2018deep,su2019robust} show promising result on several time series tasks, however, their performance on CPS data needs to be further studied. 

In this work, we propose a novel density-based time series anomaly detection framework for CPS called Neural System Identification and Bayesian Filtering (NSIBF). Specifically, we leverage an end-to-end trained neural network with a specialized architecture to capture the dynamics of CPS via a state-space model. In the detection phase, a Bayesian filtering algorithm is naturally applied on top of the ``identified" state-space model to track the uncertainty of the hidden states of the system and estimate the likelihood of observed sensor measurements over time. NSIBF is a combination of neural networks and the traditional state estimation method for CPS. As a result, NSIBF benefits from both: the generality and ability to capture highly nonlinear system dynamics; and the robustness to process and sensor noises in CPS. Extensive qualitative and quantitative experiments on a synthetic and three real-world CPS datasets show that our method achieves superior performance compared with the state-of-the-art time series anomaly detection methods. We summarize the main contributions of our paper as follows: 
\begin{itemize}
    \item We propose a novel time series anomaly detection method based on neural network-based system identification and Bayesian filtering that is robust to process and sensor noises in CPS.
    \item We studied the performance of various start-of-the-art time series anomaly detection methods on real-world CPS datasets.
    \item Through both qualitative and quantitative experiments, we show that our proposed method consistently outperforms the start-of-the-art methods on anomaly detection for CPS.
    \item We publish our data and code in a GitHub repository\footnote{https://github.com/NSIBF/NSIBF} for better reproducibility of our experimental results.
\end{itemize}

The rest of this paper is structured as follows. We briefly introduce the commonly used anomaly detection method for CPS based on state estimation in the next section. In Section~\ref{sec:nsibf} our proposed method NSIBF is presented. Section~\ref{sec:qualiexperiments} and \ref{sec:quantexperiments} present the qualitative and quantitative experiments, respectively. Related work is discussed in Section~\ref{sec:relatedwork}. Lastly, Section~\ref{sec:conclusion} draws final conclusions.

\section{Background}
\label{sec:background}
\begin{figure}
\begin{center}
\centerline{\includegraphics[width=\linewidth]{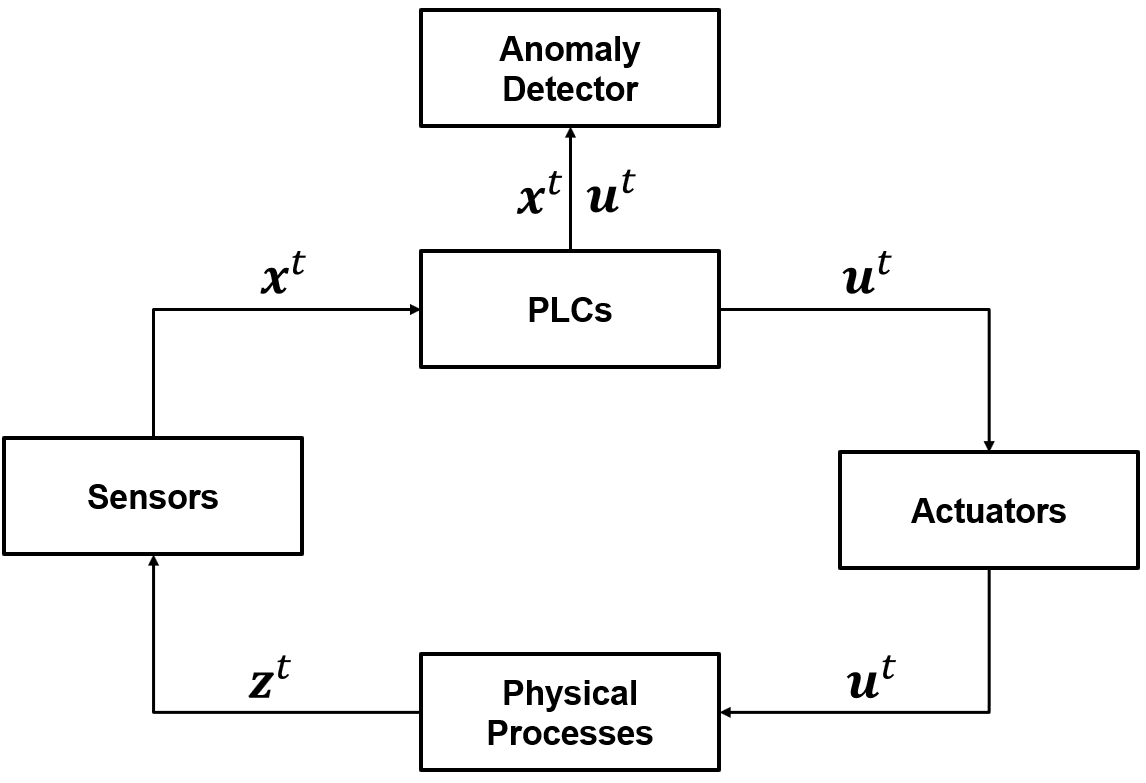}}
\caption{Anomaly detection architecture in CPS.}
\label{fig:cps}
\end{center}
\end{figure}

In a typical CPS, the physical processes are directly controlled by a network consisting of three types of field devices called sensors, actuators and Programmable Logical Controllers (PLCs) \cite{lee2017introduction}. Specifically, the sensors are devices which convert physical parameters into electronic measurements; actuators are devices which convert control commands into physical state changes (e.g., open or close a valve); based on measurements received from sensors, PLCs send control commands to actuators to activate physical state changes. To protect CPS from system failures, the physical evolution of system states are monitored via the sensor measurements and actuator states at discrete time steps. A commonly used anomaly detection architecture for CPS is illustrated in Figure~\ref{fig:cps}. Concretely, let $\mathbf{x}^t$ be the measurements from the sensors at time $t$, $\mathbf{u}^t$ be the states of actuators at time $t$, the dynamics of CPS is commonly represented by a state space model as follows:
\begin{eqnarray}
\mathbf{z}^{t} &=&  F \mathbf{z}^{t-1} + B \mathbf{u}^{t-1} + \boldsymbol{\epsilon}^t \\
\mathbf{x}^{t} &=&  H \mathbf{z}^{t} + \boldsymbol{\varepsilon}^t
\end{eqnarray}where $\mathbf{z}^{t}$ is the hidden state variables (e.g., the real temperature of liquids), $F$, $B$, $H$ are the state transition, control and measurement matrices which define the dynamics of the system, $\boldsymbol{\epsilon}^t$ and $\boldsymbol{\varepsilon}^t$ are vectors of noise for state transition and sensor measurements with a random process with zero mean. On top of the state space model, Bayesian filters such as Kalman filter \cite{kalman1960new} and its variants \cite{chui2017kalman} are commonly used to estimate the joint probability distribution of the state variables recursively over time. To detect unexpected behaviors, we need to use $\mathbf{x}^{t-1}$ and $\mathbf{u}^{t-1}$ to obtain the prior state estimation for $\mathbf{z}^{t}$, and then predict the corresponding distribution for $\mathbf{x}^{t}$. An alert is then raised when the residual error between observed measurements and the predicted mean above a threshold or the likelihood of observed measurements below a threshold \cite{giraldo2018survey}.

Despite many successful applications of the above state estimation technique in the past decades, there are some key limitations which restrict its applicability for anomaly detection of CPS in the IoT era. First of all, the step to building up the mathematical models of the process dynamics for establishing the state-space model is called ``system identification" \cite{ljung1999system}. This step is generally very challenging in practice as it requires heavy prior knowledge of the system dynamics. Furthermore, for security, privacy and other technical issues, it is often impossible for anomaly detection modules to access the control logic of the system \cite{humayed2017cyber}. In these cases, the anomaly detection model shall be able to detect unexpected system behaviors with only a partial part of the signals being accessible. Most importantly, with the growing scale and complexity of modern CPS, traditional state space models defined by domain knowledge may not be able to capture the complex dynamics of such systems any more.

\section{Neural System Identification and Bayesian Filtering}
\label{sec:nsibf}
In this section our proposed system identification method for CPS based on a specialized neural network architecture, and the corresponding Bayesian filtering method for robust time series anomaly detection are presented.

\subsection{Neural System Identification}

\begin{figure*}
\begin{center}
\centerline{\includegraphics[width=\linewidth]{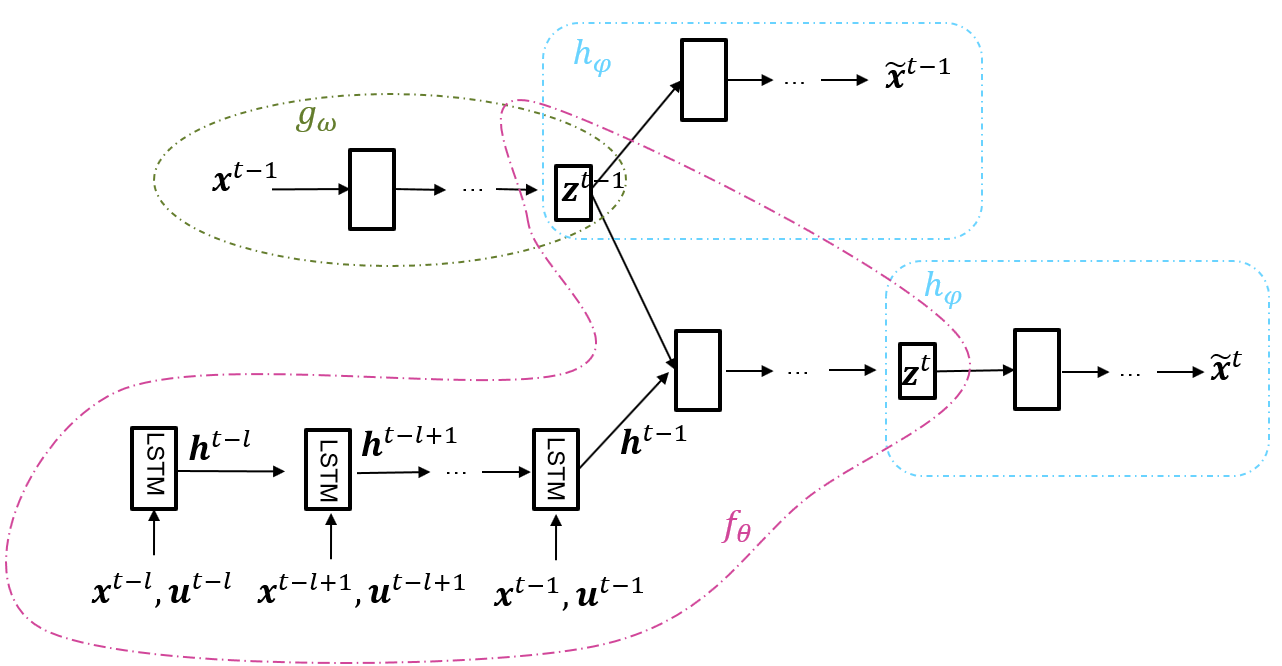}}
\caption{Proposed neural network architecture for system identification in CPS.}
\label{fig:schema}
\end{center}
\end{figure*}

Without loss of generality, let $\mathbf{x}^t$ and $\mathbf{u}^t$ be the vectors for the monitored sensor measurements and actuator states of a CPS at a discrete time point $t$, we propose the neural network architecture in Figure~\ref{fig:schema} to capture the dynamics of the system. Specifically, our proposed architecture consists of three subnets which are called the $g$ net, $f$ net and $h$ net: 
\begin{itemize}
\item The $g$ net, parameterized by $\omega$, is a feed-forward network which takes the sensor measurements at time point $t-1$ as input, and encodes the measurements into a lower dimensional hidden state vector $\mathbf{z}^{t-1}$. 
\item The $f$ net, parameterized by $\theta$, takes the sensor measurements and actuator states in a sliding window of length $l$ as input, and encodes them into a hidden vector $\mathbf{h}^{t-1}$ using recurrent layers such as LSTM \cite{hochreiter1997long}. The vector $\mathbf{h}^{t-1}$ provides the time-series context for hidden state transition. Specifically, together with $\mathbf{h}^{t-1}$, the $f$ net also takes the hidden state vector $\mathbf{z}^{t-1}$ as input, and then leverages dense layers to predict the hidden state vector at the next time point $\mathbf{z}^t$.
\item The $h$ net, parameterized by $\varphi$, is a feed-forward network which takes a hidden state vector as input, and decodes the hidden state vector to the corresponding sensor measurements. Note that the two $h$ nets in the architecture are identical meaning that they share the same parameters and weights.
\end{itemize}
Overall, we denote the whole network model as $F_{\omega,\theta,\varphi}$. The network has two inputs: a vector $\mathbf{x}^{t-1}$ which is the sensor measurements at time point $t-1$, and a sequence $(\mathbf{x},\mathbf{u})^{t-l:t-1}$ which contains the sensor measurements and actuator states in the sliding window of the past $l$ time points. There are also two outputs for the network: $\widetilde{\mathbf{x}}^{t-1}$ is the reconstructed vector of sensor measurements at time point $t-1$, $\tilde{\mathbf{x}}^{t}$ is the predicted sensor measurements at time point $t$.

$F_{\omega,\theta,\varphi}$ is trained in an end-to-end fashion. For anomaly detection purpose, the model shall be trained with CPS datasets with only normal data. Specifically, consider a CPS dataset with only normal data, we split the dataset into a training set $D_{tr}$ and a validation set $D_{val}$. Assume there are $T$ time points in $D_{tr}$, the loss function of the model is calculated as follows:
\begin{eqnarray}
L(\omega,\theta,\varphi) &=& \sum_{t=l}^{T} w_1 ||\mathbf{x}^{t-1}-\widetilde{\mathbf{x}}^{t-1}||_2^2 + w_2 ||\mathbf{x}^{t}-\widetilde{\mathbf{x}}^{t}||_2^2 \nonumber \\
&\quad&+ w_3 ||\mathbf{z}^{t}-\mathbf{z}^{t-1}||_2^2 
\end{eqnarray}in which the first two terms of $L$ are the reconstruction and prediction errors of sensor measurements, the third term is a factor to make consecutive hidden state vectors be close to each other for capturing the temporal
smoothness exists among system states. $w_1,w_2,w_3$ are hyperparameters representing the weights of the three terms.

After the model has been trained using stochastic gradient descent algorithms such as Adam \cite{kingma2014adam}, we represent the dynamics of the CPS by the following dynamical state-space model:
\begin{eqnarray}
\mathbf{z}^{t} &=&  f_{\theta} \big(\mathbf{z}^{t-1},(\mathbf{x},\mathbf{u})^{t-l:t-1} \big) + \mathbf{e}_{f}^t \\
\mathbf{x}^{t} &=&  h_{\varphi}(\mathbf{z}^{t}) + \mathbf{e}_{h}^t
\end{eqnarray}where $\mathbf{e}_{f}^t$ and $\mathbf{e}_{h}^t$ are vectors of the hidden state prediction error and the sensor measurement reconstruction error at time point $t$ respectively. Furthermore, we assume $\mathbf{e}_{f}^t$ and $\mathbf{e}_{h}^t$ follow multivariate Gaussian distribution with zero means and covariance matrices $Q$ and $R$ respectively. Specifically, $Q$ is the covariance matrix for hidden state prediction errors empirically evaluated based on the time points in $D_{val}$, and the hidden state prediction error at each time point in $D_{val}$ is calculated as follows:
\begin{equation}
\mathbf{e}_{f}^t = g_{\omega}(\mathbf{x}^{t})-f_{\theta} \big(g_{\omega}(\mathbf{x}^{t-1}), (\mathbf{x},\mathbf{u})^{t-l:t-1} \big)
\end{equation}$R$ is the covariance matrix for sensor measurement reconstruction errors which is also empirically computed based on the time points of $D_{val}$, and the sensor measurement reconstruction error at each time point is calculated as follows:
\begin{equation}
\mathbf{e}_{h}^t =\mathbf{x}^{t}-h_{\varphi}\big(g_{\omega}(\mathbf{x}^{t}) \big)
\end{equation}

Notably the above dynamical state-space model is similar to the one as illustrated in Section~\ref{sec:background}. The differences are that we replace the state transition and control matrices by the $f$ net; the measurement matrix by the $h$ net; and the process and sensor measurement noises by hidden state prediction and sensor measurement reconstruction errors, respectively.

\subsection{Bayesian Filtering and Anomaly Detection}
On top of the dynamical state-space model constructed by neural system identification, in the detection phase we keep track of the probability distribution of the hidden states of the CPS via a Bayesian filtering method, and anomalies are detected by estimating the likelihood of observed sensor measurements over time. Our method is inspired by the unscented Kalman filters \cite{julier2004unscented} which are commonly used for Bayesian state estimation for nonlinear state-space models. Compared with other nonlinear filtering methods such as particle filters \cite{kitagawa1996monte,arulampalam2002tutorial} and extended Kalman filters \cite{schmidt1966application,ribeiro2004kalman}, unscented Kalman filters often achieve a better balance between accuracy and efficiency \cite{labbekalman}.

Concretely, assume the hidden states of the system follow a multivariate Gaussian distribution, let $m$ denote the dimension of the hidden states, we use $\bar{\mathbf{z}}^t \in \mathbb{R}^{m}$ and $P^t \in \mathbb{R}^{m \times m}$ to denote the mean vector and the covariance matrix of the distribution at time $t$. We initialize $\bar{\mathbf{z}}^0=g_{\omega}(\mathbf{x}^{0})$, $P^0=\epsilon I_m$, where $\epsilon$ is a value close to zero. Then, we estimate $\bar{\mathbf{z}}^t$ and $P^t$, and conduct anomaly detection following the two steps as described below recursively over time.

\subsubsection{The prediction step}
In this step, we predict the mean and covariance of hidden state distribution at time $t$ according to known $\bar{\mathbf{z}}^{t-1}$ and $P^{t-1}$. Concretely, we first generate a set of sigma points $Z \in \mathbb{R}^{n \times m}$ and their corresponding weights $\mathbf{w}^m \in \mathbb{R}^{n}$ and $\mathbf{w}^c \in \mathbb{R}^{n}$ for mean and covariance functions respectively to efficiently represent the hidden state distribution at time $t-1$, where $n$ is the number of generated sigma points. These sigma points are smartly selected using the Julier sigma function as introduced in \cite{julier2002scaled} such that:
\begin{equation}
Z,\mathbf{w}^m,\mathbf{w}^c=\text{Sigma Function}(\bar{\mathbf{z}}^{t-1},P^{t-1}) 
\end{equation}The details of the sigma function is omitted here as it is not the focus of the paper. Then, we pass the selected sigma points through the $f$ net to generate a set of predicted state vectors at time $t$, such that
\begin{eqnarray}
Y=f_{\theta} \big(Z, (\mathbf{x},\mathbf{u})^{t-l:t-1} \big)   
\end{eqnarray}
After that, we predict the mean and covariance of the state distribution at time $t$ by the unscented transform function as follows:
\begin{eqnarray}
\hat{\mathbf{z}}^{t} &=& \sum_i  \mathbf{w}^m_i   Y_i \\
\hat{P}^{t} &=& \sum_i  \mathbf{w}^c_i   (Y_i-\hat{\mathbf{z}}^{t})(Y_i-\hat{\mathbf{z}}^{t})^T + Q
\end{eqnarray}$\hat{\mathbf{z}}^{t}$ and $\hat{P}^{t}$ are also treated as the prior mean and covariance of the hidden state distribution at time $t$.   

\subsubsection{The update and anomaly detection step}
In this step, given the observed sensor measurements, we conduct anomaly detection and update the posterior mean and covariance for the hidden state distribution at time $t$. Concretely, we firstly decode the set of predicted state vectors $Y$ at time $t$ to their corresponding sensor measurements using the $h$ net:
\begin{equation}
X=h_{\varphi}(Y)   
\end{equation}
Again, we predict mean and covariance for the sensor measurements at time $t$ through the unscented transform function:
\begin{eqnarray}
\boldsymbol{\mu} &=& \sum_i  \mathbf{w}^m_i   X_i \\
\Sigma &=& \sum_i  \mathbf{w}^c_i   (X_i-\boldsymbol{\mu})(X_i-\boldsymbol{\mu})^T  + R
\end{eqnarray}
Then, we derive an anomaly score for the observed measurements $\mathbf{x}^{t}$ by computing the Mahalanobis distance \cite{mclachlan1999mahalanobis} between $\mathbf{x}^{t}$ and its predicted distribution as follows:
\begin{equation}
    \text{anomaly score} = \sqrt{(\mathbf{x}^{t}-\boldsymbol{\mu})^T \Sigma^{-1} (\mathbf{x}^{t}-\boldsymbol{\mu})} 
\end{equation}A larger anomaly score indicates it is more likely for $\mathbf{x}^{t}$ to be an anomaly. In practice, an anomaly threshold can be tuned by methods such as setting an acceptable false alarm rate or mean time between false alarms.

To obtain the posterior mean and covariance for the state distribution at time $t$, we calculate the Kalman gain $K$ and then update $\bar{\mathbf{z}}^{t}$ and $P^{t}$ by fusing the prior estimation and the observed measurements under the Kalman framework:
\begin{eqnarray}
K &=& [\sum_i \mathbf{w}^c_i (Y_i-\hat{\mathbf{z}}^{t})(X_i-\boldsymbol{\mu})^T ]\Sigma^{-1}   \\
\bar{\mathbf{z}}^{t} &=& \hat{\mathbf{z}}^{t} + K(\mathbf{x}^{t}-\boldsymbol{\mu}) \\
P^{t} &=& \hat{P}^{t} - K \Sigma K^T
\end{eqnarray}

\subsection{Algorithm and Implementation}
To sum up, we give our method for time series anomaly detection for CPS in Algorithm~\ref{alg:realtime}. The method is implemented using the TensorFlow \cite{abadi2016tensorflow} and FilterPy \cite{labbe2018filterpy} open source libraries.

It is worth noting that although our method is described for point-wise detection, in practice we can also conduct sequence-level detection by stacking the sensor measurements in several consecutive time points into a vector form as the input for $g$ net. In this way the $\mathbf{x}^{t}$ in our proposed method becomes the stacked sensor measurements in consecutive time points. As a result, some temporal features of the measurements are also encoded in the learned hidden states and the anomalies are detected in the sequence level.
\begin{algorithm}
  \caption{Neural system identification and Bayesian filtering for anomaly detection}
  \begin{algorithmic}[1]
  	\REQUIRE A CPS dataset without anomalies
  	\STATE  \textbf{The training phase:}
  	\STATE Split the dataset into a training set $D_{tr}$ and a validation set $D_{val}$.
  	\STATE Train the neural network $F_{\omega,\theta,\varphi}$ on $D_{tr}$.
  	\STATE Estimate the covariance matrices $Q$ and $R$ using $D_{val}$.
  	\STATE \textbf{The detection phase:}
  	\STATE{Initialize $\bar{\mathbf{z}}^0=g_{\omega}(\mathbf{x}^{0})$, $P^0=\epsilon I_m$} 
  	\LOOP 
  	\STATE{Predict $\hat{\mathbf{z}}^{t}$ and $\hat{P}^{t}$, the prior mean and covariance of the state distribution at time $t$ by Equations (8-11).}
  	\STATE{Predict $\boldsymbol{\mu}$ and $\Sigma$, the mean and covariance of the sensor measurements at time $t$ by Equations (12-14).}
  	\STATE{Obtain $\mathbf{x}^t$ and $\mathbf{u}^t$, the sensor measurements and actuator states at the current time point.} 
  	\STATE{Compute the anomaly score at time $t$ by Equation (15).}
  	\STATE{Update $\bar{\mathbf{z}}^{t}$ and $P^{t}$, the posterior mean and covariance of hidden states at time $t$ by Equations (16-18).}
  	\ENDLOOP
  \end{algorithmic}
   \label{alg:realtime}
\end{algorithm}

\section{Qualitative Experiments}
\label{sec:qualiexperiments}
In our qualitative experiment, we test whether our method benefits from the Bayesian filtering approach for anomaly detection compared with merely residual error-based detection. For this, we consider a simple illustrative CPS example which simulates a noisy sine wave as follows:
\begin{eqnarray}
t \% 30=0 &\longrightarrow & u^t = 9-u^t \\
z^{t} &=& \sin({t/u^t}) + \epsilon^t \\
x^{t} &=& 2*z^{t} + \varepsilon^t
\end{eqnarray}
where $t=1,2,...,N$, $z^t$ is the hidden state of the system, $u^t$ is the state of the actuator to control the frequency of the sinusoid, $x^t$ is the observed sensor measurement; $\epsilon^t \sim \mathcal{N}(0,0.1^2)$ is the process noise, $\varepsilon^t \sim \mathcal{N}(0,0.2^2)$ is the measurement noise. With $u^0=3$, we simulate the system for 10000 time points to generate the training data. Furthermore, we simulate the system for another 10000 time points to generate the testing data in which anomalies are injected. Specifically, anomalies are injected consecutively for 100 time points in every 1000 time points. During the anomalous periods, we set $\epsilon^t \sim \mathcal{N}(0,0.6^2)$.

In our experiment, we use the data in the first 7500 time points of the training data to train our proposed neural network for system identification, the data in the remaining 2500 time points are used as the validation set. Furthermore, we conduct sequence-level anomaly detection by which we stack the sensor measurements in 31 consecutive time points (to cover at least one cycle of actuator state change) into a vector and then compress them to a two dimensional hidden state using the $g$ net. Then, the $f$ net with $l=62$ (each sliding window consists of $62$ time points) is used to predict the hidden state for the next 31 time points. Lastly, the $h$ net is used to reconstruct the sensor measurements in the current 31 time points as well as to decode the measurements in the next 31 time points. 

\begin{figure*}
\centering
\subfloat[]{\includegraphics[width=.32\textwidth]{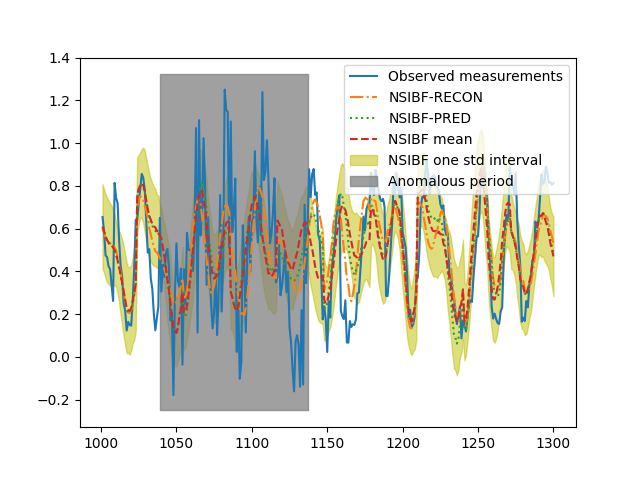}%
\label{fig:simtrace}}
\hfil
\subfloat[]{\includegraphics[width=.32\textwidth]{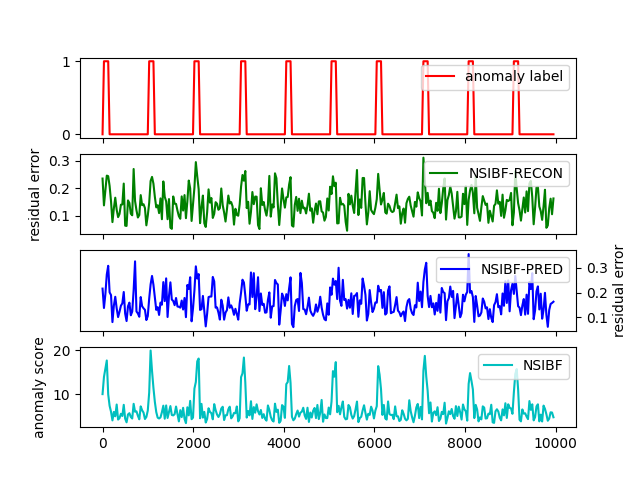}%
\label{fig:anomalyscore}}
\hfil
\subfloat[]{\includegraphics[width=.32\textwidth]{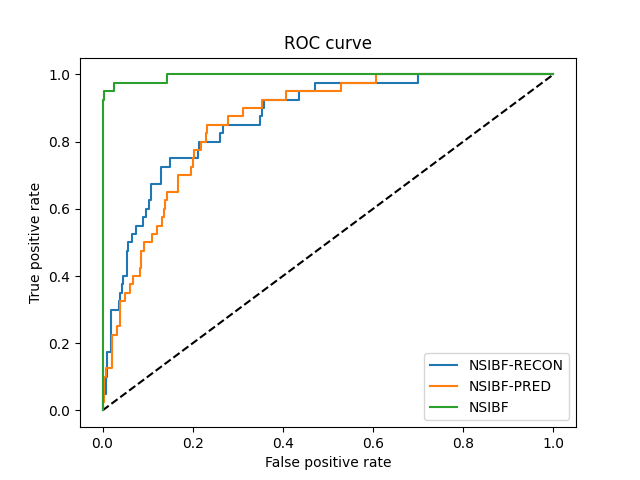}
\label{fig:roc}}
\caption{(a): the observed measurements and generated measurements by NSIBF-RECON, NSIBF-PRED and NSIBF in one anomalous period; (b): the trace of anomaly scores generated by NSIBF and residual errors generated by NSIBF-RECON and NSIBF-PRED; (c): the ROC curve for NSIBF-RECON, NSIBF-PRED and NSIBF for anomaly detection of the simulated CPS example in the qualitative experiment.}
\label{fig:sim}
\end{figure*}

As can be noticed, using our trained neural network for system identification $F_{\omega,\theta,\varphi}$, we can have three methods for anomaly detection: 1) The first method is to leverage Bayesian filtering as described in Section~\ref{sec:nsibf}, we denote this method as NSIBF; 2) The second method is to directly use the residual errors between the reconstructed measurements generated by $F_{\omega,\theta,\varphi}$ and the observed measurements to detect anomalies, we denote this method as NSIBF-RECON; 3) The third method is to directly use the residual errors between the predicted measurements generated by $F_{\omega,\theta,\varphi}$ and the observed measurements to detect anomalies, we denote this method as NSIBF-PRED. We use the above three methods to detect the anomalies in the testing data. We illustrate the observed measurements and the generated measurements by the three methods in one anomalous period in Figure~\ref{fig:simtrace}. Furthermore, the anomaly scores generated by NSIBF as well as the residual errors generated by NSIBF-RECON and NSIBF-PRED for the whole testing phase are shown in Figure~\ref{fig:anomalyscore}. The corresponding receiver operating characteristic (ROC) curves for the anomaly detection performances of the three methods are given in Figure~\ref{fig:roc}. As can be seen from Figure~\ref{fig:anomalyscore} and \ref{fig:roc}, it is harder to detect the injected anomalies by using the residual errors. However, using the anomaly scores generated by NSIBF, we can easily detect those anomalies. We believe there are two reasons: first, density-based method is more capable than residual error-based method to detect anomalies in noisy measurements; second, the reconstructed and predicted measurements by NSIBF-RECON and NSIBF-PRED are more likely to overfit the noises. But NSIBF with a underlying mechanism to track the uncertainty of the hidden states by fusing the state prediction and the observed measurements via Bayesian filtering is more robust to such noises. Thus we can see that measurements generated by NSIBF can still have a relatively stable pattern during the anomalous period as illustrated in Figure~\ref{fig:simtrace}. 

\section{Quantitative Experiments}
\label{sec:quantexperiments}
In our quantitative experiments, we apply our method for anomaly detection in three real-world CPS datasets and compare the results with various state-of-the-art anomaly detection methods.

\subsection{Datasets}
We consider the following three real-world CPS datasets:
\begin{itemize}
\item PUMP: This dataset is collected from a water pump system of a small town. The data is collected every minute for five months.
\item WADI: This dataset is collected from a fully operational physical testbed that represents a scaled-down version of a real urban water distribution system \cite{ahmed2017wadi}. The data is collected every second for 16 days. In our experiments, we downsampled it to a data point in every five seconds and the data in the last day is ignored as they have  different distributions with previous 15 days due to change of operational mode. 
\item SWAT: This dataset is collected from a CPS testbed that is a scaled down water treatment plant producing filtered water \cite{goh2016dataset}. The data is collected every second for 11 days. In our experiments, we downsampled it to a data point in every five seconds. 
\end{itemize}More details of the benchmark datasets are given in Table~\ref{tab:datasets}. Note that all the datasets are labelled by domain experts and anomalies only exist in the test data.

\begin{table}
\caption{The percentage of anomalies within the test data and the details of benchmark CPS datasets.}
\label{tab:datasets}
\begin{center}
\begin{sc}
\begin{tabular}{lccccc}
\toprule
&  Train & Test &  Num. & Num. & Anomalies \\
&  size & size &  sensors & actuators & ($\%$) \\
PUMP&  76901 & 143401 &  44 & 0 & 10.05 \\
WADI&  241921 & 15701 &  67 & 26 & 7.09 \\
SWAT&  99360 & 89984 &  25 & 26 & 11.99 \\
\bottomrule
\end{tabular}
\end{sc}
\end{center}
\end{table}

\begin{table*}
\caption{Comparison of the performance metrics of anomaly detection methods on benchmark CPS datasets.}
\label{tab:pe}
\begin{center}
\begin{sc}
\begin{tabular}{lccc|ccc|ccc}
\toprule
\multirow{2}{*}{Model} &  \multicolumn{3}{c}{PUMP} &  \multicolumn{3}{c}{WADI} &  \multicolumn{3}{c}{SWAT} \\
& Pre & Rec  &F1  & Pre & Rec & F1  & Pre & Rec & F1 \\
\midrule
Isolation Forest    & 0.977& 0.582 & 0.729  & 0.826 & 0.772 &0.798 & 0.975& 0.754  & 0.850 \\
Sparse-AE    & 0.798 &  0.737 & 0.767 & 0.769 & 0.771&0.770   & 0.999 & 0.666  & 0.799\\
EncDec-AD    & 0.438 & 0.796 & 0.565   & 0.589 & 0.887 & 0.708  & 0.945 & 0.620  & 0.748\\
LSTM-Pred    & 0.925 &  0.581 & 0.714   & 0.620 & 0.876& 0.726 & 0.996 & 0.686  & 0.812\\
DAGMM    & 0.931 & 0.798 & 0.860  & 0.886 & 0.772 & 0.825  & 0.946  &  0.747  & 0.835\\
OmniAnomaly    & 0.937& 0.840& $\mathbf{0.886}$  &  0.846 & 0.893  & $\mathbf{0.869}$ & 0.979 & 0.757  & $\mathbf{0.854}$ \\
USAD    &0.984 &0.582  & 0.732  & 0.806 & 0.879 & 0.841  & 0.987& 0.740 & 0.846\\
\hline
NSIBF-recon    & 0.859 & 0.798 & 0.827   & 0.814 & 0.495  & 0.615 & 0.999& 0.667 & 0.800\\
NSIBF-pred    & 0.897 & 0.798 & 0.845  & 0.685 & 0.876 & 0.769 & 0.982& 0.697 & 0.815\\
NSIBF & 0.936 & 0.889 & $\mathbf{0.912}$  & 0.915 & 0.887 & $\mathbf{0.901}$  & 0.982 & 0.863 & $\mathbf{0.919}$\\
\bottomrule
\end{tabular}
\end{sc}
\end{center}
\end{table*}

\begin{figure*}
\centering
\subfloat[]{\includegraphics[width=.32\textwidth]{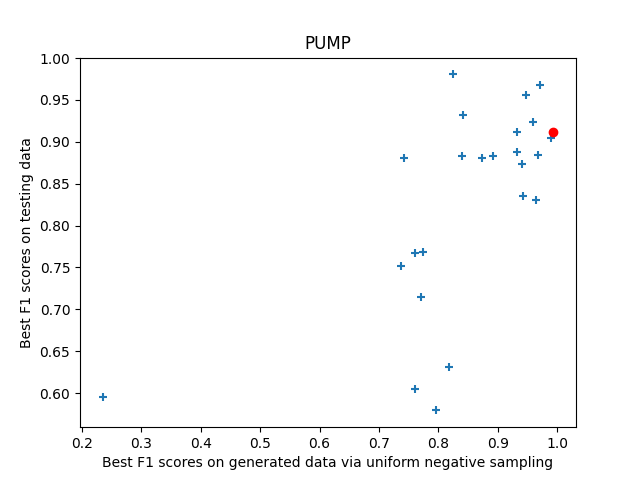}%
\label{fig:pump_hpo}}
\hfil
\subfloat[]{\includegraphics[width=.32\textwidth]{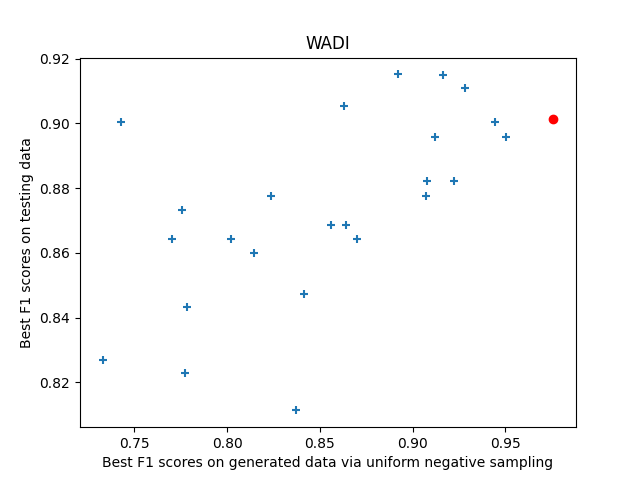}%
\label{fig:wadi_hpo}}
\hfil
\subfloat[]{\includegraphics[width=.32\textwidth]{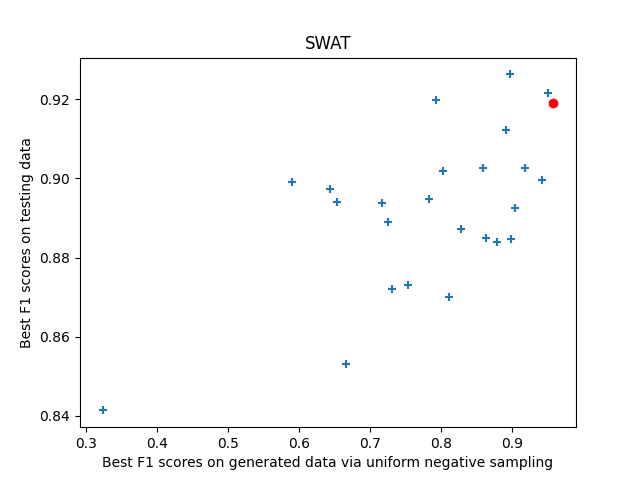}
\label{fig:swat_hpo}}
\caption{The best F1 scores on the generated datasets via uniform negative sampling during hyperparameter tuning, and the best F1 scores if the corresponding models are used for anomaly detection on the testing data for the three benchmark datasets. For each benchmark dataset, the red circle is selected as the metric for comparing with baselines in Table~\ref{tab:pe}.}
\label{fig:hpo}
\end{figure*}

\subsection{Baseline Anomaly Detection Methods}
We select several state-of-the-art anomaly detection methods as baselines.
\begin{itemize}
\item Isolation Forest: Isolation Forest \cite{liu2008isolation} is a popular anomaly detection algorithm that works on the principle of isolating anomalies using tree-based structures.
\item Sparse-AE: Sparse-AE \cite{ng2011sparse} is a residual error-based anomaly detection algorithm using an autoencoder with sparse hidden embeddings as a reconstruction model.
\item EncDec-AD: EncDec-AD \cite{malhotra2016lstm} is a residual error-based time series anomaly detection algorithm using an LSTM-based encoder-decoder as a reconstruction model. 
\item LSTM-PRED: LSTM-PRED \cite{goh2017anomaly} is a residual error-based time series anomaly detection algorithm using an LSTM-based regressor as a predictive model.
\item DAGMM: DAGMM \cite{zong2018deep} is a density-based anomaly detection algorithm using a deep generative model that
assumes the Gaussian mixture prior in the latent space to estimate the likelihood of input samples.
\item OmniAnomaly: OmniAnomaly \cite{su2019robust} is a density-based time series anomaly detection algorithm that employs a deep generative model with stochastic variable connection and planar normalizing flow to describe non-Gaussian distributions of latent space, and uses the reconstruction probabilities of input samples as anomaly scores.
\item USAD: USAD \cite{audibert2020usad} is a residual error-based time series anomaly detection algorithm that employs an autoencoder as a reconstruction model. Particularly, the autoencoder is adversarial trained to amplify the reconstruction errors for anomalies.
\end{itemize}
Note that since Isolation Forest, Sparse-AE and DAGMM are not anomaly detection methods for time series, we stack sensor measurements and actuator states in multiple consecutive time points to form a vector as the inputs for these models such that they can also capture the temporal dependence of signals in the benchmark datasets. Specifically, we stack signals of five time points for the PUMP dataset, and 12 time points for the WADI and SWAT datasets. With respect to NSIBF, we also stack the sensor measurements in five consecutive time points for the PUMP dataset and 12 time points for the WADI and SWAT datasets into vectors as the input for $g$ net. The weights $w_1,w_2,w_3$ in the loss function of Equation (3) are set to $0.45$, $0.45$ and $0.1$ for all the experiments (including the qualitative one) as we find it consistently achieves a good performance. For all datasets, we split the train data to a training set and a validation set with the ratio 3:1. To demonstrate the importance of Bayesian filtering, we also include NSIBF-RECON and NSIBF-PRED as baselines.

\subsection{Uniform Negative Sampling for Guiding Hyperparameter Tuning}
To effectively train the neural network for system identification in NSIBF, there are quite a few hyperparameters such as the dimension of the hidden state vector, the number of hidden layers and the hidden dimensions for the three subnets. We can simply use the loss function in Equation (3) as the criterion to tune hyperparameters, however it is not optimal for anomaly detection. On the other hand, in practice it is generally impossible to have a dataset with sufficient labelled anomalies to test our model's anomaly detection performance. Therefore, we propose to use the uniform negative sampling algorithm as introduced in \cite{sipple2020interpretable} to generate anomalous data from the validation set $D_{val}$ that only contains normal data. Different from \cite{sipple2020interpretable} where the uniform negative sampling algorithm is used to generate anomalous data for training classification models for anomaly detection, we utilize the generated anomalous data to guide the process of hyperparameter tuning of our model.

Specifically, we generate datasets by injecting anomalous data into $D_{val}$. Let $M$ be the dimension of $\mathbf{x}^t$, $N$ be the number of time points in $D_{val}$, for each dimension $d \leq M$, we define $lim_d = [min(U_d)-\delta,max(U_d)+\delta] $ as a range bounded by the extrema of the normal samples $U_d$ extended by a conservative positive length $\delta$ that extends $lim_d$ beyond the normal space. With a given sample ratio $r_s \in [0,1]$, we randomly select $r_s \times N$ time points from $D_{val}$. For each selected time point $t$, we replace $\mathbf{x}_t$ by a vector whose value is uniformly sampled from $lim_d$ for each $d \leq D$. Concretely, in our experiments, we generate 20 datasets with uniform negative sampling in which $r_s=[0,0.05,0.1,...,0.95]$ and $\delta=0.05$ (after normalization). We use the randomised grid search algorithm to tune the hyperparameters of NSIBF, the model with the best overall anomaly detection performance (normal data are treated as outliers in the evaluation since most sequences will contain anomalies) on the 20 generated datasets is selected for anomaly detection on the testing data. The values and ranges for the hyperparameters of NSIBF are given in the appendix.

\subsection{Evaluation Metrics and Results}
As discussed in \cite{xu2018unsupervised}, for time series anomaly detection tasks, people generally do not care about the point-wise metrics. Instead, point-adjusted metrics are favored in which a contiguous anomalous segment is considered as successfully detected as long as any point inside the segment is classified as anomaly. Moreover, in practice it is more important to have an excellent F1 score at a certain threshold than to just have generally good performance on most thresholds. As a result, following the evaluation methods in \cite{su2019robust}, for the performance of a specific method on a dataset, we enumerate all possible anomaly thresholds to search for the best point-adjusted F1 score as the main evaluation metric. 

The best F1 scores and their corresponding precision and recall for each method on all the benchmark datasets are reported in Table~\ref{tab:pe}. Furthermore, we also report the best F1 scores on the generated datasets via uniform negative sampling during hyperparameter tuning, and the best F1 scores if the corresponding models are used for anomaly detection on the testing data for the three benchmark datasets in Figure~\ref{fig:hpo}. As can be seen from Figure~\ref{fig:hpo}, using uniform negative sampling for guiding hyperparameter tuning of NSIBF can generally provide a good model for anomaly detection even without real anomalies. By selecting the model with the best performance on the generated datasets via uniform negative sampling, we simulate the scenario in practice where the model (hyperparameter setting) with the best performance on testing data is generally unknown in advance. Importantly, from Table~\ref{tab:pe} we notice that NSIBF still achieves $2.9\%$ improvement at the F1 score on the PUMP dataset, $3.7\%$ on the WADI dataset and $7.6\%$ on the SWAT dataset compared with the second best methods. Furthermore, NSIBF also significantly outperforms NSIBF-RECON and NSIBF-PRED on all the three datasets. These results clearly demonstrate the superiority of using a neural identified state-space model and Bayesian filtering to track the uncertainty of hidden states as the basis for anomaly detection in CPS signals with unknown amount of process and sensor measurement noises over time.

\section{Related Work}
\label{sec:relatedwork}
Deep learning-based time series anomaly detection in CPS has become an active research area in the recent years \cite{luo2020deep}. Till now, most of the methods focus on residual error-based detection. For example, recurrent neural networks (RNNs) are widely applied to capture the temporal dependency of time-series data in CPS \cite{feng2017multi,hundman2018detecting,tariq2019detecting,kwon2019rnn}. Autoencoders are also commonly used to capture the correlation of different sensors \cite{canizo2019multi,al2020unsupervised}. Prediction errors and reconstruction errors are used as anomaly scores. We have shown that NSIBF which is a density-based detection method tailored for CPS, can achieve better anomaly detection performance than the commonly used residual error-based methods. There are also many other semi-data-driven methods for anomaly detection in CPS. Examples are those utilize invariant rules \cite{chen2018learning,feng2019systematic}, sensor and process noise fingerprints \cite{ahmed2018noise} in specific systems. NSIBF has better generality and lower implementing difficulty compared with those methods as it is purely data-driven and does not require any case-specific domain knowledge.

To avoid the need for explicit model specification and improve the ability to capture complex time series dynamics, combining state-space models with deep neural networks has been proposed before. For example, the close relationship between RNNs and state space models that both use an internal hidden state to drive forecasts and observations is discussed in \cite{chung2015recurrent}. Applications of recurrent variational autoencoder architectures have been investigated for constructing non-linear state space models \cite{krishnan2015deep,karl2016deep,krishnan2017structured} and learning disentangled representations \cite{fraccaro2017disentangled}. In \cite{rangapuram2018deep,lim2019recurrent,salinas2020deepar}, RNN-based deep state space models are proposed for probabilistic time series forecasting. Different from previous approaches which implicitly condense both the model specification and Bayesian filtering steps into the learned neural network, we explicitly decouple the two steps in NSIBF. In this way, the state transition (process) noises and sensor measurement noises that naturally exist in CPS are captured in our framework and this strategy can provide more accurate and consistent temporal information in noisy time series, making NSIBF tailored for anomaly detection in CPS data.

\section{Conclusion}
\label{sec:conclusion}
In this paper, we introduce a novel method, which we call the Neural System Identification and Bayesian Filtering (NSIBF), to learn system dynamics and conduct time series anomaly detection in CPS data. By leveraging the expressive power of neural networks and the ability of Bayesian filters to track uncertainties, our method can accurately detect anomalies in noisy sensor data from complex CPS. Based on the qualitative experiment on the simulated CPS example and the quantitative experiments on three real-world CPS datasets, the benefits of our method can be clearly seen from the improvements on the anomaly detection metrics compared with various start-of-the-art anomaly detection methods. Importantly, as our method is highly generalizable and easy to implement, we see great potential for application of NSIBF to a wide range of use-cases in the IoT applications where CPS become more ubiquitous and commoditized.

\bibliographystyle{ACM-Reference-Format}
\bibliography{kdd}
\clearpage
\appendix

\section{Baseline Implementations}
Isolation Forest comes from the scikit-learn implementation\footnote{https://scikit-learn.org/stable/modules/generated/sklearn.ensemble.IsolationForest.html}. DAGMM comes from a Tensorflow implementation in the Github repository\footnote{https://github.com/tnakae/DAGMM}. OmniAnomaly comes from a Tensorflow implementation in the Github repository\footnote{https://github.com/NetManAIOps/OmniAnomaly}. USAD comes from a PyTorch implementation in the Github repository\footnote{https://github.com/robustml-eurecom/usad}. Sparse-AE, EncDec-AD and LSTM-Pred are implemented by us using Tensorflow. All experiments are run on a Linux machine with 64 GiB memory, 8 4.2GHz Intel Cores and a GTX 1080 GPU.

\section{Hyperparameter Configurations of NSIBF}
\begin{table}[H]
\begin{center}
\begin{sc}
\begin{tabular}{lcc}
\toprule
Hyperparameter & Dataset & Value or Range \\
\midrule
$w_1$ &  All & 0.45 \\
$w_2$ &  All & 0.45 \\
$w_3$ &  All & 0.1 \\
$l$ &  PUMP & 15 \\
$l$ &  WADI & 36 \\
$l$ &  SWAT & 36 \\
Num. dimensions for $\mathbf{z}_t$ &  PUMP & $[1,120]$ \\
Num. dimensions for $\mathbf{z}_t$ &  WADI & $[1,200]$ \\
Num. dimensions for $\mathbf{z}_t$ &  SWAT & $[1,75]$ \\
Num. hidden layers for $g$ net &  All & [1,3] \\
Num. hidden layers for $h$ net &  All & [1,3] \\
Num. dense layers for $f$ net &  All & [1,3] \\
Num. LSTM layers for $f$ net &  All & [1,3] \\
Num. dimensions in hidden layers & All & [32,256] \\
Training epochs & All & 100 \\
\bottomrule
\end{tabular}
\end{sc}
\end{center}
\end{table}

\end{document}